\documentclass[10pt,twocolumn,letterpaper]{article}

\usepackage{cvpr}      
\usepackage{multirow}
\definecolor{cvprblue}{rgb}{0.21,0.49,0.74}
\usepackage[pagebackref,breaklinks,colorlinks,allcolors=cvprblue]{hyperref}

\def\paperID{40935} 
\def\confName{CVPR}
\def\confYear{2026}

\title{Stability-Driven Motion Generation for Object-Guided Human-Human Co-Manipulation}

\author{Jiahao Xu$^{1}$\hspace{5mm} Xiaohan Yuan$^{2}$ \hspace{5mm} Xingchen Wu$^{1}$\hspace{5mm} Chongyang Xu$^{3}$\hspace{5mm} Kun Li$^{1}$\hspace{5mm}  Buzhen Huang$^{1}$\footnotemark[1]\\%
\\
$^1$Tianjin University \hspace{1mm}
$^2$National University of Singapore \hspace{1mm} 
$^3$Sichuan University \hspace{1mm}\\
\\
\vspace{-13mm}
}

\begin{document}

\twocolumn[{%
\renewcommand\twocolumn[2][]{#1}%
\maketitle
\thispagestyle{empty}
\vspace{-7mm}
\begin{center}
   \centering
   \includegraphics[width=0.95\textwidth]{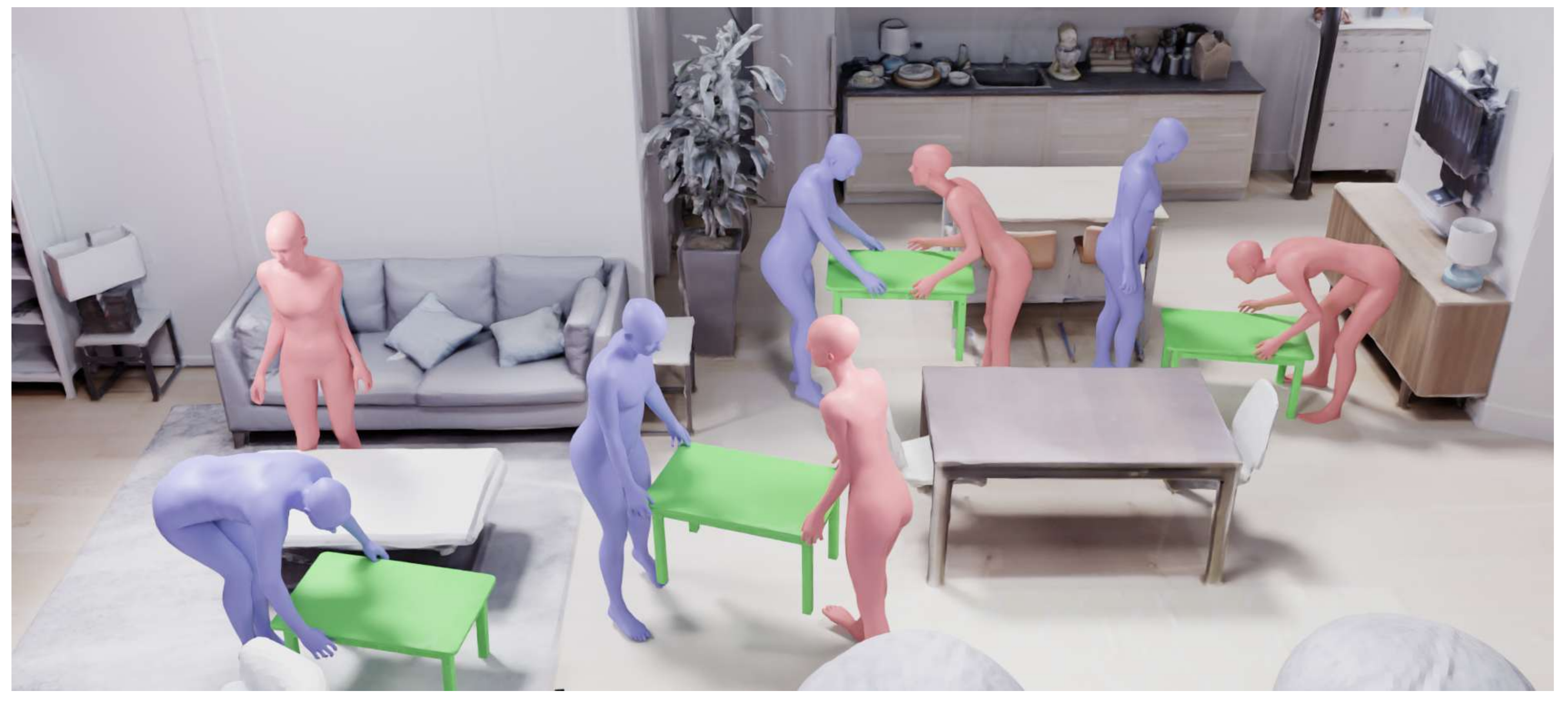}
   \vspace{-3mm}
   \captionof{figure}{Given an object mesh and its trajectory~(green), our method generates coordinated motions that are consistent with the trajectory while remaining natural and physically plausible for co-manipulation.}
   \label{fig:teaser}
\end{center}
}]

\footnotetext[1]{Corresponding author: Buzhen Huang, hbz@tju.edu.cn.}

\begin{abstract}

Co-manipulation requires multiple humans to synchronize their motions with a shared object while ensuring reasonable interactions, maintaining natural poses, and preserving stable states. However, most existing motion generation approaches are designed for single-character scenarios or fail to account for payload-induced dynamics. In this work, we propose a flow-matching framework that ensures the generated co-manipulation motions align with the intended goals while maintaining naturalness and effectiveness. Specifically, we first introduce a generative model that derives explicit manipulation strategies from the object’s affordance and spatial configuration, which guide the motion flow toward successful manipulation. To improve motion quality, we then design an adversarial interaction prior that promotes natural individual poses and realistic inter-person interactions during co-manipulation. In addition, we also incorporate a stability-driven simulation into the flow matching process, which refines unstable interaction states through sampling-based optimization and directly adjusts the vector field regression to promote more effective manipulation. The experimental results demonstrate that our method achieves higher contact accuracy, lower penetration, and better distributional fidelity compared to state-of-the-art human-object interaction baselines. The code is available at \url{https://github.com/boycehbz/StaCOM}.

\end{abstract}

\section{Introduction}
\label{sec:intro}
Modeling coordinated human motion in interactive environments is a key problem in computer graphics, virtual reality, and robotics. The challenge intensifies in scenarios where multiple agents interact with shared objects (\eg, two-person lifting or collaborative manipulation), since each agent must adapt to both the object's motion and the partner's behavior. Such human-human co-manipulation involves tightly coupled triadic interactions, which require synchronized motion between humans and objects and physical feasibility during joint manipulation.

However, most existing motion generation methods are tailored to single-person scenarios, where an individual moves in response to static environments or predefined object trajectories~\cite{li2023object,tevet2022human}. These models lack the mechanisms for inter-agent communication and mutual adaptation, making them unsuitable for collaborative tasks in co-manipulation. On the other hand, recent multi-person motion generation methods \cite{liang2024intergen,li2024interdance} typically focus on social or dance-like interactions without involving object manipulation. Directly adapting these frameworks to co-manipulation cannot ensure reasonable physical dynamics and coherent human–object coordination. Consequently, developing motion generation models that can capture both inter-agent coordination and realistic human–object dynamics for multi-person co-manipulation remains an open research question.

We observe that real-world co-manipulation motions need to satisfy three key aspects: \textbf{1) Intention}: the manipulation strategy should be determined based on the object's shape, affordance, and goal state to achieve the desired manipulation. \textbf{2) Naturalness}: human motion should be natural and responsive to the partner's actions. \textbf{3) Effectiveness}: the transport process should be stable and comply with physical laws. Based on these observations, we propose a flow-matching framework that integrates object affordances, motion priors, and physics-based feedback into the motion generation process. Guided by these principles, our model generates co-manipulation motions that better satisfy intention, naturalness, and physical plausibility.

Specifically, we adopt flow-matching~\cite{lipman2022flow} as our basic framework, which learns a continuous vector field to map noise into clean motion data under object trajectory guidance. We further employ BPS representation~\cite{prokudin2019efficient} to enhance the model's perception of dynamic object pose and shape. Building upon this, we introduce an affordance-informed manipulation strategy to guide motion generation during inference. Since humans can produce the same object motion through diverse movements, this strategy generates graspability fields conditioned on object affordances, and the resulting contact anchors serve as explicit gradient cues that attract hands toward high-probability regions on the object surface, yielding geometrically consistent yet diverse manipulation plans. Nonetheless, relying solely on the manipulation strategy often results in stiff or poorly synchronized poses. We therefore develop an adversarial interaction prior that scores dual-human motion naturalness by analyzing joint rotations, inter-agent timing, and role symmetry. In contrast to previous discriminators~\cite{kanazawa2018end,goel2023humans}, this prior focuses on collaborative cues and penalizes misaligned reactions, so the flow receives gradients that preserve coordinated behaviors. In addition, we incorporate a stability-driven simulation to inject physics-based feedback, where a sampling-based formulation refines unstable poses and steers the generator toward motions that maintain grasp stability and limit payload drift. The main contributions of this work are summarized as follows:

\begin{itemize}
  \item We propose a flow-matching framework to consider intention, naturalness, and effectiveness principles to generate physically plausible and socially coordinated human co-manipulations.

  \item We introduce an affordance-informed manipulation strategy and an interaction prior that ensure natural interactions while producing plausible manipulations.

  \item We consider co-manipulation motion with a focus on stability and propose a sampling-based simulation to jointly refine the interaction and motion.

\end{itemize}

\section{Related Work}\label{sec:relatedwork}
\begin{figure*}
    \centering
    \includegraphics[width=1.0\textwidth, keepaspectratio]{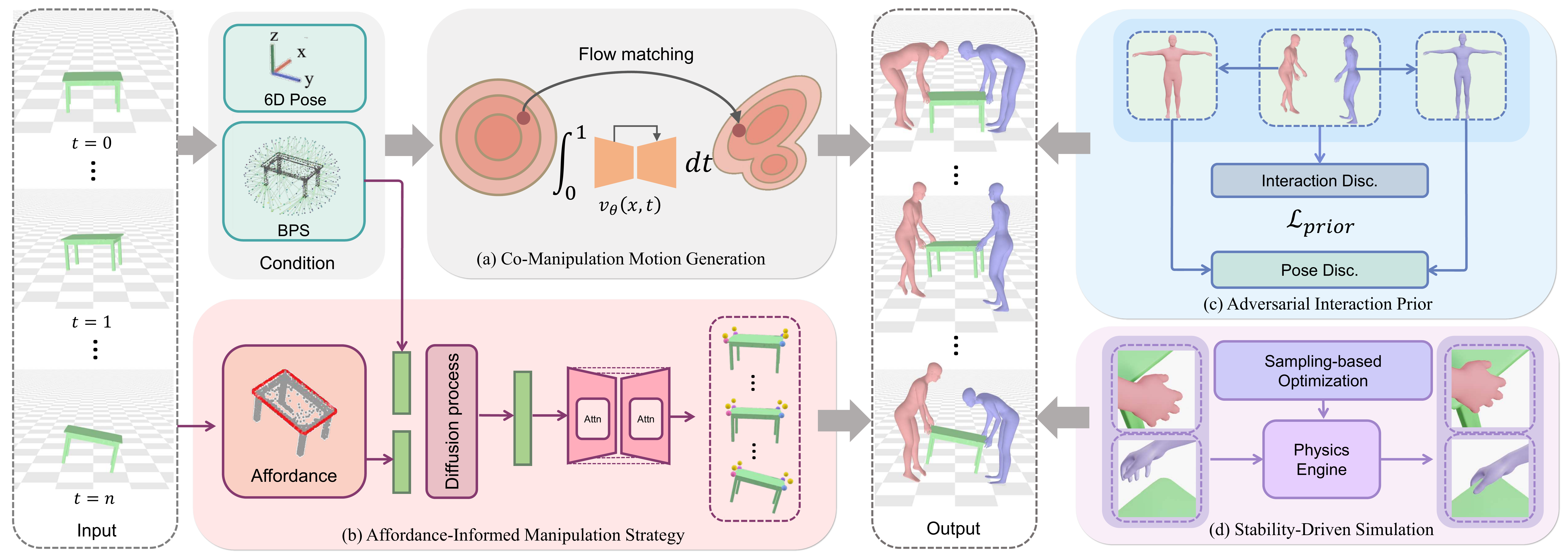}
    \vspace{-5mm}
    \caption{\textbf{Overview.} Given an input object trajectory, our method generates co-manipulation motions conditioned on object 6D poses and their BPS features (a). To ensure that the motions are consistent with the object trajectory, an affordance-informed manipulation strategy (b) is introduced to produce explicit contact signals as flow guidance. Building on this design, we further propose an adversarial interaction prior (c) and a stability-driven simulation (d) to enhance motion quality. Note that the contact strategy~(b), flow matching~(a), and interaction prior~(c) are trained separately in advance, while all components are executed jointly at inference time.}
    \label{fig:method_pipeline}
    \vspace{-5mm}
\end{figure*}

\paragraph{Human-Object Interaction Generation.}
Human-object interaction (HOI) generation has evolved from contact-aware diffusion to affordance-driven reasoning, yet most methods still emphasize single-actor scenes~\cite{xu2025interact,wu2025human,pi2023hierarchical,li2024controllable,liu2024mimicking,wu2025hoi}. InterDiff~\cite{xu2023interdiff} and CG-HOI~\cite{diller2024cg} inject contact objectives into diffusion pipelines, whereas OMOMO~\cite{li2023object}, NIFTY~\cite{kulkarni2024nifty}, and NAP~\cite{lei2023nap} reconstruct interactions from object trajectories or articulated priors to provide semantic control. With the availability of advanced human–object–human datasets~\cite{liu2025core4d,xu2025perceiving}, recent works have begun to explore collaborative manipulation. OnlineHOI~\cite{ji2025onlinehoi} introduces a Mamba-based network for generating human motions conditioned on object geometry. SyncDiff~\cite{he2025syncdiff} promotes phase-aligned multi-body motion, while COLLAGE~\cite{daiya2025collage} samples cooperative contacts for dual agents. However, these methods rely solely on stochastic denoising without incorporating explicit physical feedback. CooHOI~\cite{gao2024coohoi} achieves physically plausible cooperative human–object interactions through reinforcement learning, but its policy fails to generalize across different objects and tasks. In contrast to prior works, our approach advances human co-manipulation motion generation by integrating deterministic flow matching with affordance-guided gradients and adversarial coordination priors. A generalizable sampling-based simulation module is further introduced to improve manipulation stability.

\paragraph{Multi-person Interaction Generation.}
Motion generation for multi-person interaction focuses on modeling dynamic interactions between individuals~\cite{zhu2025react,ji2025towards,xu2025multi,ota2025pino,chang2025large}. However, existing methods have long inference times that are not applicable to real-time generation of scenarios. To improve the model inference speed, Li~\etal~\cite{li2024two} introduced a factorization strategy, which decomposes the joint probability of interaction motion into three independent distributions and effectively captures the geometric and topological relationships in the interaction between two people. Some works generate actions conditional on textual descriptions~\cite{xu2024regennet,wu2025text2interact}, combining boosted VAE models and a diffusion-based framework to capture spatio-temporal dependencies to enhance the modeling of two-player interactions ~\cite{javed2024intermask,li2025two,liang2024intergen,xu2024inter,liu2025ponimator}. Other approaches integrate physical constraints such as contact optimization~\cite{li2024interdance} and collision detection~\cite{huang2024closely,wang2024intercontrol} to alleviate human penetration and improve realism. While these methods can generate higher-quality two-person interaction actions, they do not scale to more complex multi-person or even interaction actions with objects. Directly applying generic multi-person motion generation frameworks to co-manipulation scenarios may lead to severe artifacts.

\paragraph{Physics-based Motion Generation.}
Physics-grounded motion synthesis blends data-driven priors with simulators or analytic constraints to ensure physical plausibility~\cite{tevet2024closd}. Optimization-based controllers such as~\cite{huang2022neural,liu2015improving} integrate tracking objectives with physics solvers to reconstruct human motions under contact and torque limits. PhysDiff~\cite{yuan2023physdiff} augments diffusion with physics-constrained gradient guidance. Reinforcement-learned character controllers such as DeepMimic~\cite{peng2018deepmimic} and AMP~\cite{peng2021amp} further couple motion capture priors with physics-based policies to maintain balance, enforce contact stability, and suppress penetration. The technology is improved in recent years with more advanced designs like universal representation~\cite{luo2023perpetual,luo2023universal}, diverse reward functions~\cite{yuan2023learning,hassan2023synthesizing}, and motion priors~\cite{zhang2025physics,tessler2024maskedmimic,shi2024interactive,pan2025tokenhsi}. Despite these advances, existing methods primarily address single-agent locomotion or isolated human-object interactions. In contrast, we operate on dual-human co-manipulation, combining gradient-based contact objectives with simulator-in-the-loop selection to handle tightly coupled human-object dynamics without additional policy training.

\section{Method}\label{sec:Method}

We aim to generate coordinated human–human co-manipulation motions conditioned on an object’s geometry and trajectory. An overview is illustrated in \cref{fig:method_pipeline}. Our framework first employs a flow-matching model to generate coordinated human motions conditioned on the object’s configuration~(\cref{sec:flow}). Built upon this model, we introduce a diffusion-based module that generates contact strategies guided by object affordances, providing explicit cues to refine the motion flow~(\cref{sec:strategy}). We further incorporate an adversarial regularizer~(\cref{sec:prior}) and a stability-driven simulation module~(\cref{sec:simulation}) to enhance the realism and physical stability of the generated motions. The details of each component are described in the following sections.

\subsection{Interaction representation}\label{sec:Representation}
We represent each individual with SMPL-X~\cite{pavlakos2019expressive} tuple $\mathbf{x}_t^{(a)} = (\boldsymbol{\theta}_t^{(a)},  \boldsymbol{\beta}^{(a)}, \gamma_t^{(a)})$ at frame $t$, where $\boldsymbol{\theta}^{(a)}\in\mathbb{R}^{J\times 6}$ denotes joint rotations in 6D representation~\cite{zhou2019continuity}. $\gamma^{(a)}\in\mathbb{R}^3$ and $\boldsymbol{\beta}^{(a)}\in\mathbb{R}^{10}$ are global translation and shape coefficients for agent $a\in\{1,2\}$. Concatenating the two agents yields the interaction sequence $\mathbf{x}=\{(\mathbf{x}_t^{(1)},\mathbf{x}_t^{(2)})\}_{t=0}^{T}$. The object is described by its mesh $\mathcal{O}$ and rigid motion trajectory $\{(R^o_t,\mathbf{d}^o_t)\}_{t=0}^{T}$. We also employ a Basis Point Set~(BPS) descriptor $\mathbf{b}_t\in\mathbb{R}^{1024}$~\cite{prokudin2019efficient} computed from a fixed sampling of $\mathcal{O}$ to provide per-frame shape context. Contact information is encoded as $\mathcal{C}_t=\{(\mathbf{p}_{a,h}^t,\mathbf{n}_{a,h}^t,\boldsymbol{\delta}_{a,h}^t,s_{a,h}^t)\}$, where $\mathbf{p}_{a,h}^t$ and $\mathbf{n}_{a,h}^t$ are the world-space position and normal of the contact point for hand $h\in\{\text{left},\text{right}\}$ of agent $a$. $\boldsymbol{\delta}_{a,h}^t$ is the local offset relative to the mesh vertex, and $s_{a,h}^t\in\{0,1\}$ indicates contact validity. To model graspability priors for contact estimation, we also predict object affordance $\mathcal{A}=\{(\mathbf{q}_k,\alpha_k)\}$ on sampled surface points $\mathbf{q}_k\in\mathcal{O}$, where $\alpha_k$ is the affordance probability for the $k$th point.

\subsection{Co-manipulation motion generation}\label{sec:flow}
Human-human co-manipulation requires satisfying multiple criteria, including goal alignment, motion naturalness, and task effectiveness. To model such realistic and coordinated interactions, we adopt flow matching~\cite{lipman2022flow} as our foundational framework, which formulates motion generation as a velocity field regression that transports a noise sample $\mathbf{x}_0$ toward the data distribution $\mathbf{x}_1$. Flow matching enables stable likelihood-based training and avoids stochastic sampling via a deterministic vector field, which can efficiently incorporate various conditions as guidance to steer motion generation. 

We therefore propose a transformer-based flow $f_\theta$ to estimate the instantaneous interaction velocity from states $\mathbf{x}_\tau$ and conditioning $\mathbf{c}$, where $\tau$ is a continuous parameter ranging from 0 to 1. The condition $\mathbf{c}$ concatenates the object pose descriptors $\{(R^o_t, \mathbf{d}^o_t)\}$, BPS embeddings $\{\mathbf{b}\}$, and cached contact anchors before projection. The flow network predicts a velocity that transports the current state toward the data manifold. The update is written as
\begin{equation}
\mathbf{x}_{\tau+\Delta \tau} = \mathbf{x}_\tau + \Delta \tau\, f_\theta(\mathbf{x}_\tau, \tau, \mathbf{c}).
\label{eq:euler}
\end{equation}
The network performs $K$ Euler integration steps to evolve the initial noise $\mathbf{x}_0$ into the reconstructed motion. The training procedure minimizes the mean-squared flow objective:
\begin{equation}
\mathcal{L}_{\text{flow}} = \mathbb{E}_{\tau,\mathbf{x}_\tau}\Big[\big\| f_\theta(\mathbf{x}_\tau, \tau, \mathbf{c}) - (\mathbf{x}_1-\mathbf{x}_0) \big\|_2^2\Big],
\label{eq:flow}
\end{equation}
which follows the continuous flow-matching derivation. To reduce the sensitivity to outliers in articulated joints, we additionally supervise an element-wise $L_1$ loss on the decoded SMPL-X parameters,
\begin{equation}
\mathcal{L}_{\text{SMPL}} = \mathbb{E}_{\tau}\Big[\big\| \hat{\mathbf{x}}_1 - \mathbf{x}_1^{\text{gt}} \big\|_1\Big],
\label{eq:l1flow}
\end{equation}
where $\hat{\mathbf{x}}_1$ denotes the reconstructed SMPL-X parameters predicted at flow step $\tau$ and $\mathbf{x}_1^{\text{gt}}$ is the target state. This clear separation keeps the flow objective focused on estimating the continuous velocity field while the additional $L_1$ term stabilizes the articulated-body decoding. To further suppress foot-sliding artifacts, we incorporate a foot-contact loss~\cite{tevet2022human}:
\begin{equation}
\mathcal{L}_{\text{foot}} = \left\|(\mathbf{J}_{f}^{t+1} - \mathbf{J}_{f}^{t}) \cdot f^{t}\right\|_2^2,
\label{eq:foot_loss}
\end{equation}
where $\mathbf{J}_f$ denotes the 3D position of the foot joints, and $f^{t} \in \{0,1\}$ is the binary foot-contact mask at frame $t$. This loss penalizes foot displacement during contact frames, effectively suppressing foot-sliding artifacts. We optimize the weighted sum of the flow, SMPL, foot-contact, and prior objectives
\begin{equation}
\mathcal{L}_{\text{total}} = \mathcal{L}_{\text{flow}} + \mathcal{L}_{\text{SMPL}} + \mathcal{L}_{\text{foot}} + \mathcal{L}_{\text{prior}}.
\label{eq:total_loss}
\end{equation}
$\mathcal{L}_{\text{prior}}$ serves as an adversarial loss that encourages realistic and coherent motion generation, and its formulation is detailed in \cref{sec:prior}. The resulting motions are decoded to SMPL-X meshes, and the gradient-based contact refinement from \cref{sec:strategy} keeps the synthesized wrists aligned with the stored hand-object anchors for coherent bimanual manipulation.

\subsection{Manipulation strategy generation}\label{sec:strategy}

Although the flow matching network can generate plausible interactive motions conditioned on specific object information and contact points, humans can produce the same object motion with varying movements. To this end, we propose an affordance-informed contact prediction network to generate diverse strategies for a specific manipulation task. For a given object geometry, we first train a regression network~\cite{li2024laso} with dense contact annotations to predict the affordance probability $\alpha_k$ for sampled surface points. As shown in \cref{fig:method_pipeline}~(b), the predicted affordance and BPS features are used as conditions to a diffusion model to predict the contact strategy $\mathcal{C}=\{(\mathbf{p},\mathbf{n},\boldsymbol{\delta},s)\}$ from pure noises. The diffusion model is supervised with the following constraints:
\begin{equation}
\mathcal{L}_{\text{str}} = \mathcal{L}_{\text{anchor}} + \mathcal{L}_{\text{normal}} + \mathcal{L}_{\text{aff}}.
\label{eq:str_loss}
\end{equation}

The contact-anchor loss constrains the generated anchors $\hat{\mathbf{p}}$ to stay close to the ground-truth contact points $\mathbf{p}$:
\begin{equation}
\mathcal{L}_{\text{anchor}} = \frac{1}{Z_{\text{pos}}}\sum_{t,a,h} s_{a,h}^t \big\|\hat{\mathbf{p}}_{a,h}^t - \mathbf{p}_{a,h}^t\big\|_2^2,
\end{equation}
where $s$ is a binary flag indicating whether contact occurs, and $Z_{\text{pos}} = \sum_{t,a,h} s_{a,h}^t$ serves as a normalization factor over valid anchors. In parallel, a normal-alignment objective is also applied:
\begin{equation}
\mathcal{L}_{\text{normal}} = \frac{1}{Z_{\text{pos}}}\sum_{t,a,h} s_{a,h}^t \big(1 - \hat{\mathbf{n}}_{a,h}^t \cdot \mathbf{n}_{a,h}^t\big).
\end{equation}

To highlight the affordance guidance, we introduce a simple regularizer that encourages each sampled contact to lie in graspable regions suggested by the affordance predictor.
\begin{equation}
\mathcal{L}_{\text{aff}} = -\frac{1}{Z_{\text{pos}}}\sum_{t,a,h} s_{a,h}^t \log \alpha(\hat{\mathbf{p}}_{a,h}^t),
\end{equation}
Evaluating the learned affordance field $\alpha(\cdot)$ at the predicted anchor $\hat{\mathbf{p}}_{a,h}^t$ encourages the diffusion model to sample from surface zones with high graspability scores. These constraints ensure that the generated manipulation strategies remain consistent with positional, directional, and affordance cues while still allowing diverse motion variations.

The predicted contacts are then used to guide the motion flow. Specifically, we minimize a differentiable distance loss that keeps the human wrists close to the contact points during flow matching. Let $\mathcal{V}_{a,h}$ be the validity indicator for hand $h$ of agent $a$ and $\hat{\mathbf{p}}_{a,h}$ the corresponding contact anchor. Given the wrist positions $\mathbf{w}_{a,h}$ decoded from the current state, we evaluate
\begin{equation}
\mathcal{L}_{\text{contact}} = \frac{1}{Z}\sum_{a,h} \mathcal{V}_{a,h} \big\|\mathbf{w}_{a,h} - \hat{\mathbf{p}}_{a,h}\big\|_2^2,
\label{eq:contact_loss}
\end{equation}
where $Z=\sum_{a,h} \mathcal{V}_{a,h}$ normalizes over active constraints. During each Euler step, we adjust the flow prediction by the loss gradient,
\begin{equation}
\tilde{f}_\theta(\mathbf{x}_\tau) = f_\theta(\mathbf{x}_\tau) - \gamma \, \nabla_{\mathbf{x}_\tau} \mathcal{L}_{\text{contact}},
\label{eq:guided_flow}
\end{equation}
With the manipulation strategy and contact guidance, the flow matching network produces trajectories that follow the desired object motion yet remain consistent with the affordance-weighted contact plans.

\subsection{Adversarial interaction prior}\label{sec:prior}
A realistic human-human co-manipulation should reflect natural interactions. However, the motion may exhibit artifacts when relying solely on contact guidance, thereby affecting its naturalness. We therefore adopt two adversarial discriminators operating at both the individual pose and paired interaction levels to improve motion quality. As shown in \cref{fig:method_pipeline}~(c), the pose prior $\mathcal{D}_\phi^{\text{body}}$ focuses on individual poses and takes as input per-joint rotation matrices and SMPL shape coefficients. It applies $1\times1$ convolutions across the 21 joint rotation blocks to extract joint-wise realism cues, which are then fused with shape-aware MLP branches to produce realness score. In parallel, the interaction prior $\mathcal{D}_\phi^{\text{int}}$ processes the concatenation of dual-agent rotations, relative root transformations, and fused shape descriptors to capture inter-person coordination cues that cannot be inferred from individual body observations.

Specifically, ground-truth poses from datasets and generated poses are used as real and fake samples, respectively. However, since individual poses in Core4D may contain artifacts, Core4D samples are excluded from the real sample set in the individual pose prior $\mathcal{D}_\phi^{\text{body}}$. Likewise, the interaction discriminator is trained with paired trajectories from datasets as positive samples and synthesized dual-agent sequences as negative ones, encouraging $\mathcal{D}_\phi^{\text{int}}$ to focus on cooperative behaviors observed in real co-manipulation data.

Each discriminator is optimized with a non-saturating binary cross-entropy objective:
\begin{align}
\mathcal{L}_{\text{prior}}^{(k)} &= - \mathbb{E}_{(\mathbf{R},\boldsymbol{\beta}) \sim \mathcal{D}_{\text{real}}^{(k)}}\big[\log \mathcal{D}_\phi^{k}(\mathbf{R}, \boldsymbol{\beta})\big] \\
&\quad - \mathbb{E}_{(\tilde{\mathbf{R}},\tilde{\boldsymbol{\beta}}) \sim \mathcal{D}_{\text{gen}}^{(k)}}\big[\log (1 - \mathcal{D}_\phi^{k}(\tilde{\mathbf{R}}, \tilde{\boldsymbol{\beta}}))\big],
\end{align}
where $k \in \{\text{body}, \text{int}\}$ indexes the single-body and interaction priors, and $\mathbf{R}$ denotes the input of the prior. The losses sum to $\mathcal{L}_{\text{prior}} = \mathcal{L}_{\text{prior}}^{(\text{body})} + \mathcal{L}_{\text{prior}}^{(\text{int})}$ and the gradients propagate through the flow decoder to encourage realistic articulation and coordination during training.

Beyond training, the learned priors can also enhance motion naturalness during flow matching. The trained discriminators are reused as evaluators that guide the sampling process, where the predicted state is refined using the aggregated gradients $\nabla_{\mathbf{x}_\tau} \log \mathcal{D}_\phi^{k}$, \ie, gradients that encourage the sampled poses to align with realistic human motion patterns.
\begin{equation}
\tilde{f}_\theta(\mathbf{x}_\tau) = f_\theta(\mathbf{x}_\tau) + \eta \sum_{k \in \{\text{body},\text{int}\}} \nabla_{\mathbf{x}_\tau} \log \mathcal{D}_\phi^{k},
\end{equation}
where $\eta$ is a guidance weight. This correction guides the integration toward regions favored by the learned priors while preserving the base velocity field.

\begin{figure}
    \begin{center}
    \includegraphics[width=1.0\linewidth]{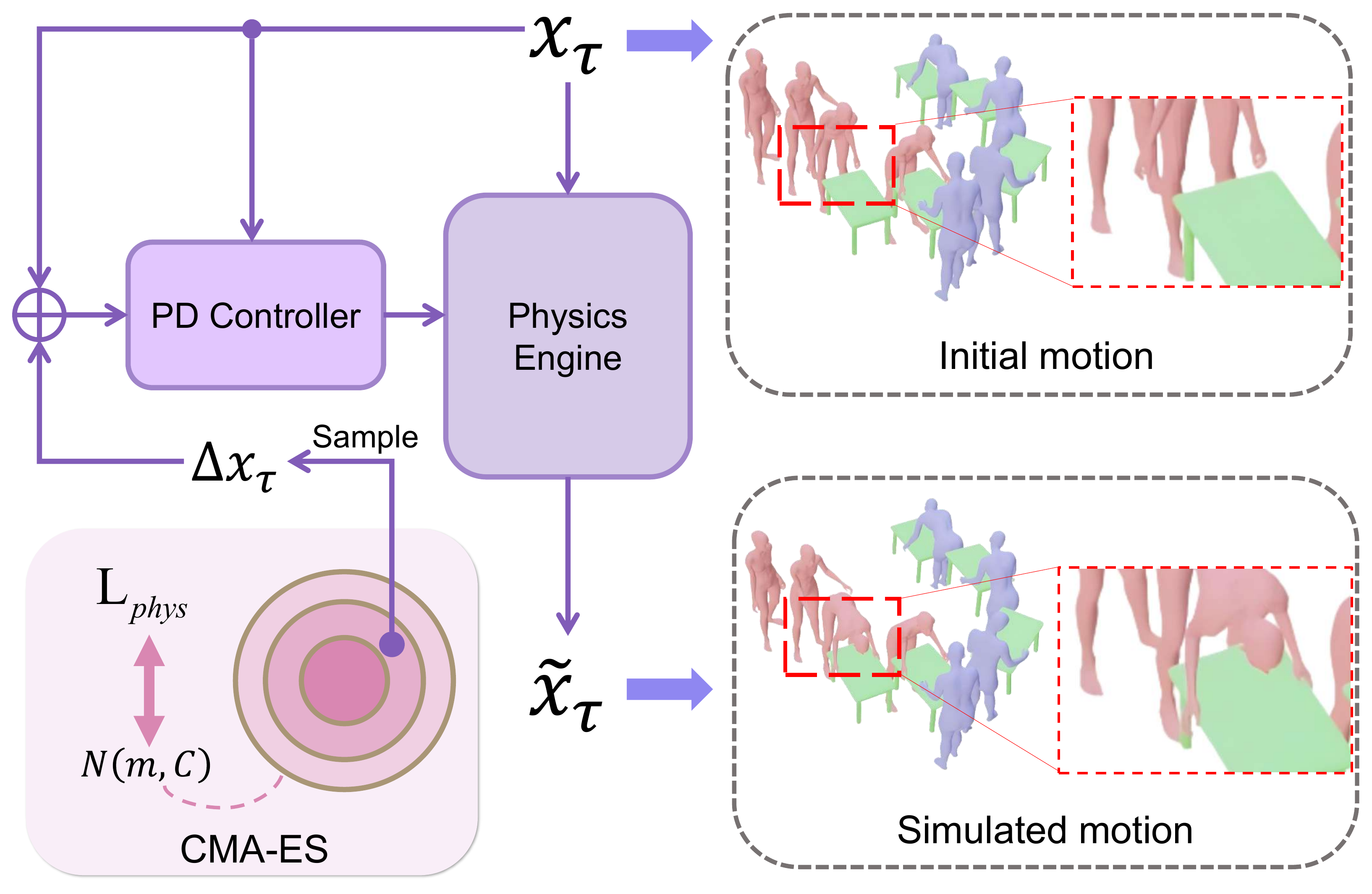}
    \end{center}
    \vspace{-5mm}
    \caption{\textbf{Stability-driven simulation pipeline.} The CMA-ES algorithm samples corrective offsets $\Delta \mathbf{x}_\tau$ for the flow-matching outputs $\mathbf{x}_\tau$. The corrected motions are then fed into the physics engine equipped with a PD controller, and the simulated results are used in the next Euler integration step.}
    \vspace{-5mm}
    \label{fig:simulation}
\end{figure}

\subsection{Stability-driven simulation}\label{sec:simulation}
Although the generated co-manipulation motions align with the input object trajectory and exhibit plausible interactions, they often suffer from severe floating and penetration artifacts between hands and manipulated objects, leading to object instability and physically implausible motion. To address this limitation, we further introduce a stability-driven simulation. Since the current RL-based policies~\cite{gao2024coohoi,pan2025tokenhsi} cannot be generalized to diverse objects and tasks, we adopt sampling-based optimization~\cite{liu2015improving,huang2022neural,gartner2022trajectory} to achieve the refinement. During the flow matching process, we execute the simulation in the penultimate integration step, and then the simulated results are directly used for the last step. 

Specifically, we convert $\mathbf{x}_\tau$ into SMPL-X parameters and instantiate humanoid models based on the SMPL-X meshes. The decoded pose and translation are used to initialize the humanoids in the physics engine. Without further correction, artifacts such as floating or penetration may cause the object to fall during simulation. Therefore, we use a proportional-derivative (PD) controller with CMA-ES algorithm~\cite{hansen2016cma} to adjust the body poses. As shown in \cref{fig:simulation}, based on the initial parameters $\mathbf{x}_\tau$, we sample a correction $\Delta \mathbf{x}_\tau$ from a multivariate normal distribution $\mathcal{N}(\boldsymbol{m}, \boldsymbol{C})$ to construct the desired targets $\bar{\mathbf{x}}_\tau = \mathbf{x}_\tau + \Delta \mathbf{x}_\tau$ for the PD controller. The distribution $\mathcal{N}(\boldsymbol{m}, \boldsymbol{C})$ is initialized as a standard normal and updated during CMA-ES optimization. During simulation, the PD controller produces joint torques to drive the humanoids toward the target poses. We then evaluate each sample with several cost functions. 
\begin{equation}
\mathcal{L}_{\text{phys}} = \mathcal{L}_{\text{sim}} + \mathcal{L}_{\text{sta}}.
\label{eq:phys_loss}
\end{equation}

The similarity loss evaluate the similarity between simulated body and object poses:
\begin{equation}
\mathcal{L}_{\text{sim}} = \|\tilde{\mathbf{x}}_\tau - \mathbf{x}_\tau\|_2^2 + \|\tilde{R}^o - R^o\|_2^2 + \|\tilde{\mathbf{d}}^o - \mathbf{d}^o\|_2^2,
\label{eq:sim_loss}
\end{equation}
where $\tilde{\mathbf{x}}_\tau$ and $\{\tilde{R}^o, \tilde{\mathbf{d}}^o\}$ are the simulated results for humanoids and object, respectively. The stability loss is formulated as:
\begin{equation}
\mathcal{L}_{\text{sta}}= \frac{\left\|\vec{f}(t)-M_o \vec{a}\right\|_2^2}{\left\|M_o \vec{g}\right\|_2^2}+\frac{\left\|\vec{\mu}(t)-I_o \vec{\alpha}\right\|_2^2}{\left\|I_o \vec{\alpha}\right\|_2^2}+e^{-m(t)},
\end{equation}
where $\vec{f}(t)$ and $\vec{\mu}(t)$ are resultant external force and torque. $M_o$ and $I_o$ are mass and inertia matrix of the object. $\vec{a}$ and $\vec{\alpha}$ are linear and angular acceleration calculated from input trajectories. $e^{-m(t)}$ is an energy regularization~\cite{peng2018deepmimic}, and $\vec{g}$ is the gravity.

We refine the sampling distribution through CMA-ES optimization and take the simulation result with the lowest cost $\tilde{\mathbf{x}}_\tau$ as the final output. The simulated trajectories are fed into the next integration step. With the simulation, this refinement leads to more physically plausible and stable co-manipulation behaviors.

With the manipulation strategy, interaction prior, and simulation, our method can finally generate intention-driven manipulation with natural and effective motions.
\section{Experiments}\label{sec:Experiments}
In this section, we first introduce the datasets and evaluation metrics used in our experiments, followed by implementation details for reproducibility. We then compare our approach with state-of-the-art methods to demonstrate its effectiveness. Finally, we perform ablation studies to analyze the contribution of each key component.

\subsection{Datasets}\label{sec:datasets}

\textbf{Core4D}~\cite{liu2025core4d} is a large-scale dataset focusing on collaborative human–object–human interactions. We extract the ground-truth object trajectories from its sequences and use them as input to our 3D motion generation framework. We adopt the official training and testing splits defined in Core4D. \textbf{Inter-X}~\cite{xu2024inter} is a large-scale dataset designed for versatile human–human interaction analysis. We utilize this dataset to enhance the quality and diversity of our motion generation model.

\begin{figure*}
    \begin{center}
    \includegraphics[width=1.0\linewidth]{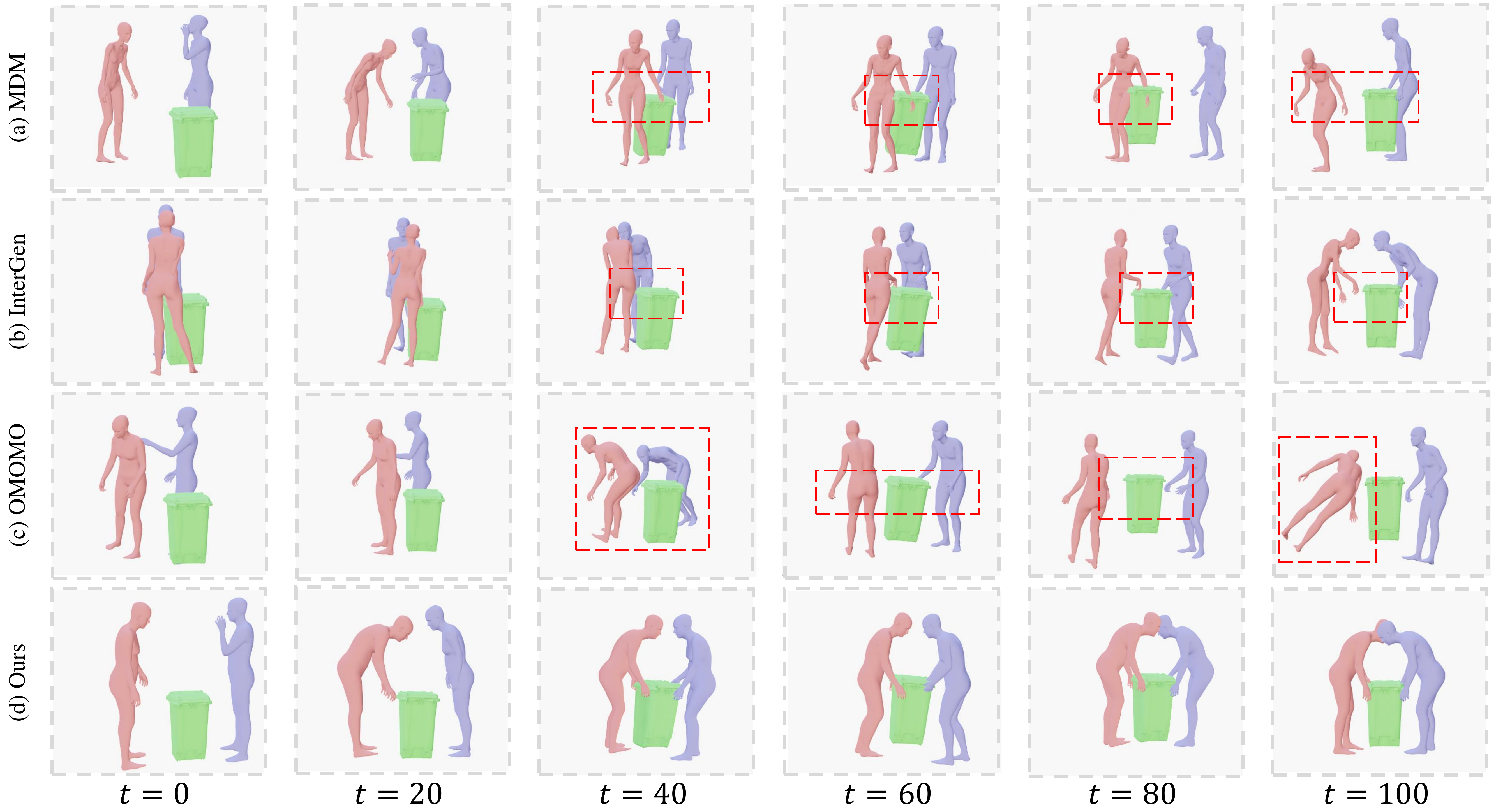}
    \end{center}
    \vspace{-5mm}
    \caption{Qualitative comparison on Core4D-S1, showing manipulations generated by ComMDM, InterGen, and OMOMO (a–c), as well as our approach (d), at key timestamps $t \in {0, 20, 40, 60, 80, 100}$. Our results (d) maintain coordinated grasps and stable payload alignment, whereas previous methods exhibit slipping contacts or delayed responses when the green object changes its pose.}
    \label{fig:comparison}
\end{figure*}

\subsection{Metrics}\label{sec:Metrics}
We adopt several metrics to evaluate motion and interaction quality. For interaction, we use interactive Distance Field (IDF)~\cite{xue2025guiding} to measure the fidelity of human-object spatial relationships via the mean-squared error between predicted and ground-truth implicit distance fields sampled around the object. In addition, \textbf{Contact Accuracy (Contact Acc.)}~\cite{liu2025core4d} evaluates how well binary hand-object contacts are reproduced using frame-level precision. \textbf{Penetration (Pene.)}~\cite{huang2024closely} quantifies physical plausibility by averaging signed penetration depths between reconstructed human meshes and the object. To evaluate motion quality, we report the \textbf{Fréchet Inception Distance (FID)} for distributional realism and \textbf{Diversity (Div.)} defined as the average pairwise distance between generated sequences.

\subsection{Implementation Details}\label{sec:details}
We train and evaluate our framework using a single NVIDIA RTX~4090 GPU with 24\,GB memory, an Intel Xeon Platinum CPU, and 90\,GB of RAM. Training follows the default configuration with a batch size of 10, a learning rate of $1\times10^{-4}$, AdamW optimization, and a cyclic cosine scheduler. Flow-matching inference uses $K=10$ Euler integration steps with a stability refinement. Physical simulation leverages PyBullet, where the physics simulator runs at 240\,Hz while the proportional-derivative controller issues targets at 60\,Hz to match the contact-conditioned motion sampling rate. The flow matching model and physics simulation take 1.19 s and 3 min, respectively, to generate a 128-frame motion sequence.

\begin{table*}[h]
    \centering
    \caption{Performance comparison on the Core4D dataset. Metrics marked with $\uparrow$ indicate higher is better, while $\downarrow$ indicates lower is better.}
    \vspace{-1mm}
    \resizebox{1.0\linewidth}{!}{
    \begin{tabular}{l|ccccc|ccccc}
    \toprule
        \multirow{2}{*}{Method}& \multicolumn{5}{c|}{Core4D-S1}  & \multicolumn{5}{c}{Core4D-S2} \\
        \cmidrule(lr){2-11}
         & IDF  $\downarrow$ & Contact Acc. $\uparrow$ & FID $\downarrow$ & Div. $\uparrow$ & Pene. $\downarrow$ & IDF $\downarrow$ & Contact Acc. $\uparrow$ & FID $\downarrow$ & Div. $\uparrow$ & Pene. $\downarrow$ \\
        \midrule
         ComMDM~\cite{shafir2023human}  & 0.41 & 0.11 & 52.5 & 1.13 & 0.19 & 0.43 & 0.12  & 49.4 & 1.13 & 0.21 \\
         OMOMO~\cite{li2023object} & 0.38 & 0.21 & 45.8 & 1.08  & 0.15 & 0.37 & 0.23 & 44.4  & 1.10 & 0.15 \\
         InterGen~\cite{liang2024intergen} & 0.47 & 0.13 & 35.4 & \textbf{1.21} & 0.11 &0.47 & 0.10 & 30.2 & \textbf{1.18} & 0.12 \\
         Ours & \textbf{0.22} & \textbf{0.44} & \textbf{25.5} & 1.15 & \textbf{0.05} & \textbf{0.20} & \textbf{0.46} & \textbf{21.6} & \textbf{1.18} & \textbf{0.06} \\
        \bottomrule
    \end{tabular}
    }
    \vspace{-4mm}
    \label{tab:core4d_comparison}
\end{table*}

\begin{figure*}
    \begin{center}
    \includegraphics[width=0.9\linewidth]{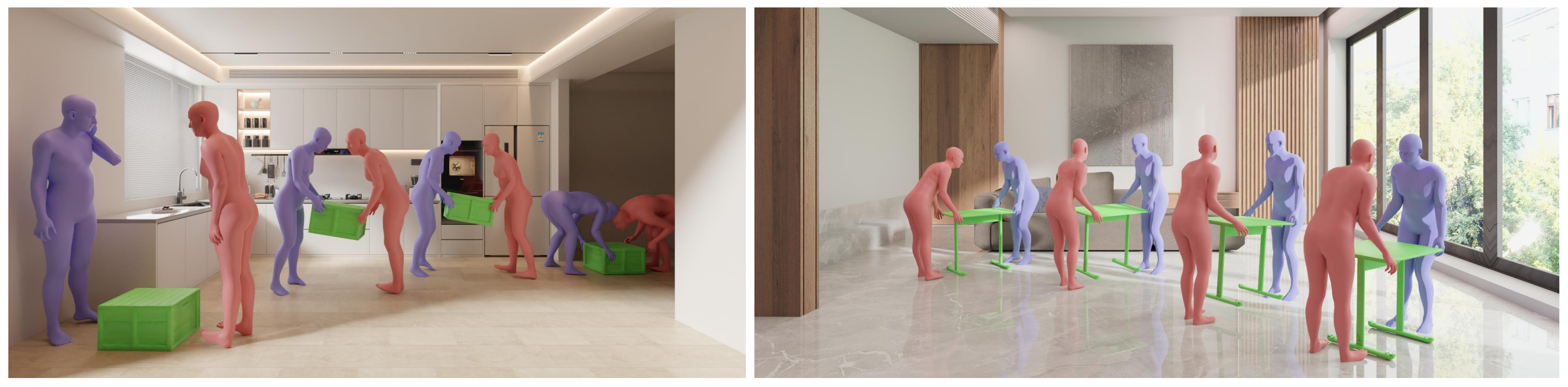}
    \end{center}
    \vspace{-6mm}
    \caption{Cooperative motions produced by our framework. The two characters remain synchronized while steering and lifting the green object along the given trajectory, exhibiting fine-grained grasp readjustments throughout the interaction.}
    \vspace{-4mm}
    \label{fig:results}
\end{figure*}

\begin{figure}
    \begin{center}
    \includegraphics[width=1.0\linewidth]{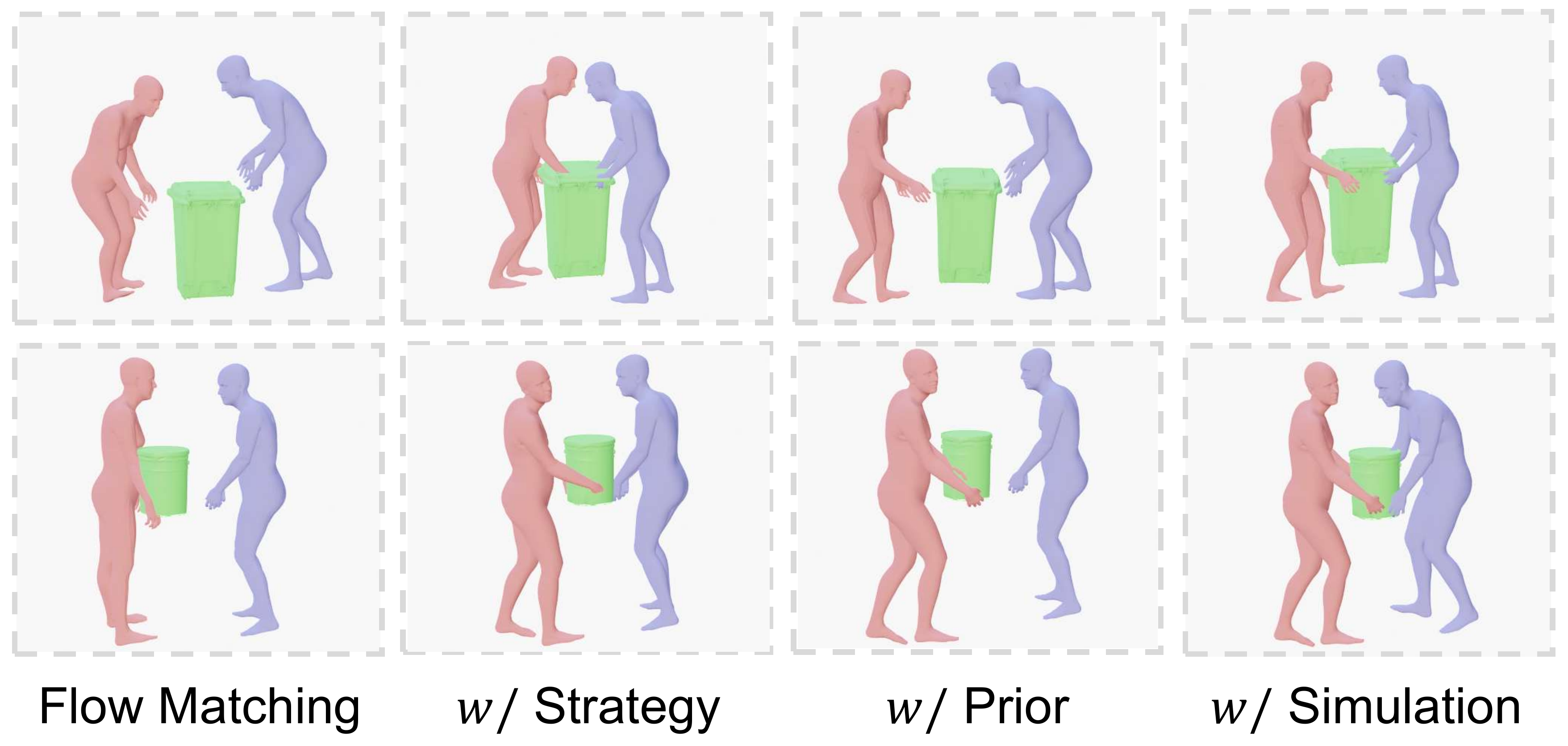}
    \end{center}
    \vspace{-5mm}
    \caption{Ablation of key components on Core4D-S1. The vanilla flow matching model fails to generate realistic interactions. Built upon this baseline, the affordance-informed strategy and interaction prior further improve motion realism and naturalness. Moreover, the physics-based simulation enhances physical plausibility.}
    \label{fig:ablation}
\end{figure}

\subsection{Comparison to State-of-the-Art Methods}\label{sec:Comparison}
Since no existing open-source human–human–object interaction methods are directly comparable to ours, we adapt several state-of-the-art human–human and human–object interaction methods as baselines. To align with our object-guided task, we modify ComMDM~\cite{shafir2023human} and InterGen~\cite{liang2024intergen} by replacing their text-conditioning inputs with our object 6D poses and BPS features. Furthermore, we extend OMOMO~\cite{li2023object}, an object-guided human–object interaction generation method, to handle dual-human scenarios.

\cref{tab:core4d_comparison} show a quantitative comparison on Core4D test sets. Since ComMDM and InterGen do not explicitly incorporate object trajectories, their performance on IDF and Contact Acc. is inferior to ours. Although OMOMO imposes contact constraints, it still exhibits unstable dual-motion coordination and frequent penetrations due to the heuristic extension from single-character to dual-human modeling. Furthermore, the affordance-informed manipulation strategy offers diverse and semantically consistent guidance, leading to improved FID and comparable diversity, which demonstrates the effectiveness of our approach in generating realistic and varied motions. Moreover, by incorporating physical simulation, our method also achieves lower penetration rates and improved physical plausibility compared to other kinematics-based baselines.

The qualitative comparison in \cref{fig:comparison} shows that when two people cooperate to move an object, both ComMDM and OMOMO suffer from orientation misalignment and fail to generate natural and coordinated interactions with the same object. Although InterGen employs a Transformer with cross-attention to improve the naturalness of dual-human actions, the lack of contact constraints between the humans and the object leads to significant hand–object misalignment. In contrast, our method, guided by affordance priors and physical simulation, maintains stable contact among both humans and the object during $t{=}40$–$100$. Moreover, benefiting from the adversarial interaction prior, our method consistently produces natural and physically plausible interactive postures throughout the manipulation process.

\begin{table}[h]
    \centering
    \caption{Ablation studies on Core4D-S1 dataset. ``Flow Matching" uses 6D poses as the trajectory without additional modules, and ``+” indicates the inclusion of the corresponding module on top of the Flow Matching baseline.}
    \vspace{-2mm}
    \resizebox{1.0\linewidth}{!}{
    \begin{tabular}{l|ccccc}
    \toprule
        Method
         & IDF  $\downarrow$ & Contact Acc. $\uparrow$ & FID $\downarrow$ & Div. $\uparrow$ & Pene. $\downarrow$ \\
        \midrule
         Flow Matching & 0.25 & 0.35 & 26.3 & 1.16 & 0.15 \\
         + BPS feature & 0.24 & 0.37 & 26.1 & 1.15 & 0.14 \\
         + Contact & 0.24 & 0.40 & 26.0 & 1.16 & 0.20 \\
         + GT Contact & 0.25 & 0.42 & 26.1 & 1.15 & 0.20 \\
         + Individual Prior & 0.22 & 0.34 & 25.5 & 1.15 & 0.16 \\
         + Interaction Prior & 0.23 & 0.35 & 25.4 & 1.16 & 0.16 \\
         + Simulation & 0.23 & 0.42 & 28.6 & 1.15 & 0.02 \\
         Ours & 0.22 & 0.44 & 25.5 & 1.15 & 0.05 \\
        \bottomrule
    \end{tabular}
    }
    \vspace{-3mm}
    \label{tab:interhuman_comparison}
\end{table}

\subsection{Ablation Study}\label{sec:Ablation}
\paragraph{Physics-based simulation.} 
The stability-driven simulation module rolls out the predicted motions using a PD controller and optimizes torques via CMA-ES to correct unstable postures before the final denoising step. Although the simulated motions are physically plausible, they may still exhibit unnatural poses due to the lack of prior knowledge~(\ie, FID). Therefore, the motions are further finetuned with the last integration step of the Flow Matching model. With this pipeline, \cref{tab:interhuman_comparison} and \cref{fig:ablation} show that our method generates realistic motions while maintaining physical plausibility. Removing the simulation leads to a notable drop in contact accuracy from 0.44 to 0.37 and a significant increase in penetration depth from 0.05 to 0.16, confirming that the physics feedback is essential for stable and accurate co-manipulation. The FID remains largely unchanged, demonstrating that the final flow-matching step successfully recovers motion naturalness after physics-based correction.
\vspace{-8mm}

\paragraph{Adversarial interaction prior.} 
The adversarial interaction prior consists of two components: an individual pose discriminator that refines local articulation, and an interaction prior that encourages coordinated full-body interactions. Compared with the vanilla Flow Matching model in \cref{tab:interhuman_comparison}, adding the individual pose prior decreases IDF from 0.25 to 0.22 and FID from 26.3 to 25.5, improving single-pose generation quality despite a temporary drop in contact accuracy. Adding the dual-agent prior subsequently increases contact accuracy to 0.35 while maintaining FID at 25.4, indicating that adversarial guidance enhances interactions without sacrificing realism.
\vspace{-4mm}

\paragraph{Manipulation Strategy.} 
The affordance module predicts dense scores on the object surface and guides the diffusion sampler to generate contact anchors that satisfy positional, normal, and availability constraints. These anchors are then used to refine hand–object alignment during motion generation. Using the predicted anchors to optimize hand–object contact increases contact accuracy from 0.35 to 0.40, while FID remains at 26.0, as shown in \cref{tab:interhuman_comparison} under ``+ Contact". This demonstrates that availability-aware anchors can significantly improve hand–object alignment.
\vspace{-2mm}

\section{Conclusion}\label{sec:Conclusion}
We propose a collaborative manipulation generation framework that conditions dual SMPLX humans on object geometry and trajectories. To ensure that the generated motion aligns with the input trajectories, we introduce an affordance-informed manipulation strategy that provides explicit contact guidance for the flow matching model. Furthermore, an adversarial interaction prior and a physics-based simulation are incorporated to further enhance motion realism and physical plausibility. With these modules, our method can generate realistic and natural human–human co-manipulation motions from given object information.

\noindent\textbf{Acknowledgements}
This work was supported by Science Fund for Distinguished Young Scholars of Tianjin under Grant 22JCJQJC00040.

{
    \small
    \bibliographystyle{ieeenat_fullname}
    \bibliography{main}
}


\end{document}